\documentclass[10pt,twocolumn,letterpaper]{article}

\usepackage{cvpr}
\usepackage{times}
\usepackage{epsfig}
\usepackage{graphicx}
\usepackage{amsmath}
\usepackage{amssymb}

\usepackage{tabularx,ragged2e}
\usepackage{booktabs,calc}
\usepackage{multirow}
\usepackage{hhline}
\usepackage{array}
\usepackage{float}
\usepackage{subcaption}
\restylefloat{table}
\usepackage{makecell}
\usepackage{color}
\usepackage{mathtools}
\usepackage{enumitem}
\usepackage[export]{adjustbox}
\mathchardef\mhyphen="2D
\usepackage[font={small},skip=0pt]{caption}

\usepackage{authblk}
\usepackage[pagebackref=true,breaklinks=true,letterpaper=true,colorlinks,bookmarks=false]{hyperref}

\DeclareMathOperator*{\argmin}{argmin}

\def\Tab{Tab\onedot}

\cvprfinalcopy % *** Uncomment this line for the final submission

\ifcvprfinal\fi

\begin{document}
\newcommand\threecycles{\mathop{3\mhyphen \text{cycles}}}
\newcommand{\KU}{\underline{K}}
\newcommand{\KI}{\overline{K}}

%\newcolumntype{Y}{>{\centering\arraybackslash}X}
%\renewcommand\tabularxcolumn[1]{>{\Centering}p{#1}}
\newcolumntype{Y}{>{\centering\arraybackslash}X}
\newcolumntype{s}{>{\hsize=.1\hsize}Y}
\newcolumntype{t}{>{\hsize=.2\hsize}Y}
\newcolumntype{u}{>{\hsize=.24\hsize}Y}

%%%%%%%%% TITLE
\title{PoseTrack: Joint Multi-Person Pose Estimation and Tracking}
\author[1]{\vspace{-6mm}Umar Iqbal}
\author[2]{Anton Milan}
\author[1]{Juergen Gall\vspace{-3mm}}
\affil[1]{Computer Vision Group, University of Bonn, Germany}
\affil[2]{Australian Centre for Visual Technologies, University of Adelaide, Australia\vspace{-3mm}}
\maketitle
%\thispagestyle{empty}

%%%%%%%%% ABSTRACT
\begin{abstract}
 In this work, we introduce the challenging problem of joint multi-person pose estimation and tracking of an unknown number of persons in unconstrained videos. Existing methods for multi-person pose estimation in images cannot be applied directly to this problem, since it also requires to solve the problem of person association over time in addition to the pose estimation for each person. We therefore propose a novel method that jointly models multi-person pose estimation and tracking in a single formulation. To this end, we represent body joint detections in a video by a spatio-temporal graph and solve an integer linear program to partition the graph into sub-graphs that correspond to plausible body pose trajectories for each person. The proposed approach implicitly handles occlusion and truncation of persons. Since the problem has not been addressed quantitatively in the literature, we introduce a challenging ``Multi-Person PoseTrack'' dataset, and also propose a completely unconstrained evaluation protocol that does not make any assumptions about the scale, size, location or the number of persons. Finally, we evaluate the proposed approach and several baseline methods on our new dataset. 

\end{abstract}

%%%%%%%%% BODY TEXT
\section{Introduction}
Human pose estimation has long been motivated for its applications in understanding human interactions, activity recognition, video surveillance and sports video analytics. The field of human pose estimation in images has progressed remarkably over the past few years. The methods have advanced from pose estimation of single pre-localized persons \cite{pishchulin2016deepcut, carreira2015human, wei2016convolutional, hu2016bottom, insafutdinov2016deepercut, newell2016eccv, bulat2016human, rafi2016bmvc} to the more challenging and realistic case of multiple, potentially overlapping and truncated persons \cite{gkioxari2014using, chen2015parsing, pishchulin2016deepcut, insafutdinov2016deepercut, Iqbal_ECCVw2016}. 
Many applications, such as mentioned before, however, aim to analyze human body motion over time.     
While there exists a notable number of works that track the pose of a single person in a video \cite{park_iccv2011, cherian_cvpr2014, dong2015iccv, ramakrishna_cvpr2013, zuffi_iccv2013, jain_accv2014, Pfister15a, charles2016cvpr, georgia2016eccv, iqbal2017FG}, multi-person human pose estimation in unconstrained videos has not been addressed in the literature. 

In this work, we address the problem of tracking the poses of multiple persons in an unconstrained setting. This means that we have to deal with large pose and scale variations, fast motions, and a varying number of persons and visible body parts due to occlusion or truncation. In contrast to previous works, we aim to solve the association of each person across the video and the pose estimation together. To this end, we build upon the recent methods for multi-person pose estimation in images~\cite{pishchulin2016deepcut, insafutdinov2016deepercut, Iqbal_ECCVw2016} that build a spatial graph based on joint proposals to estimate the pose of multiple persons in an image. In particular, we cast the problem as an optimization of a densely connected spatio-temporal graph connecting body joint candidates spatially as well as temporally. The optimization problem is formulated as a constrained Integer Linear Program (ILP) whose feasible solution partitions the graph into valid body pose trajectories for any unknown number of persons. In this way, we can handle occlusion, truncation, and temporal association within a single formulation.   

\begin{figure}[t]
  \centering
    \includegraphics[width=0.49\linewidth]{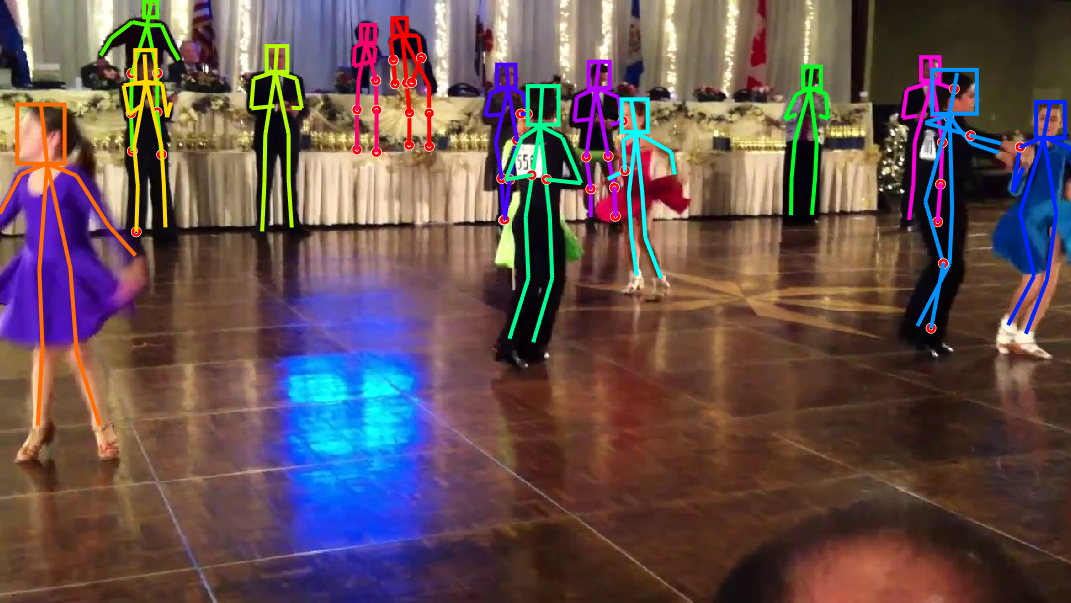} 
    \hfill
    \includegraphics[width=0.49\linewidth]{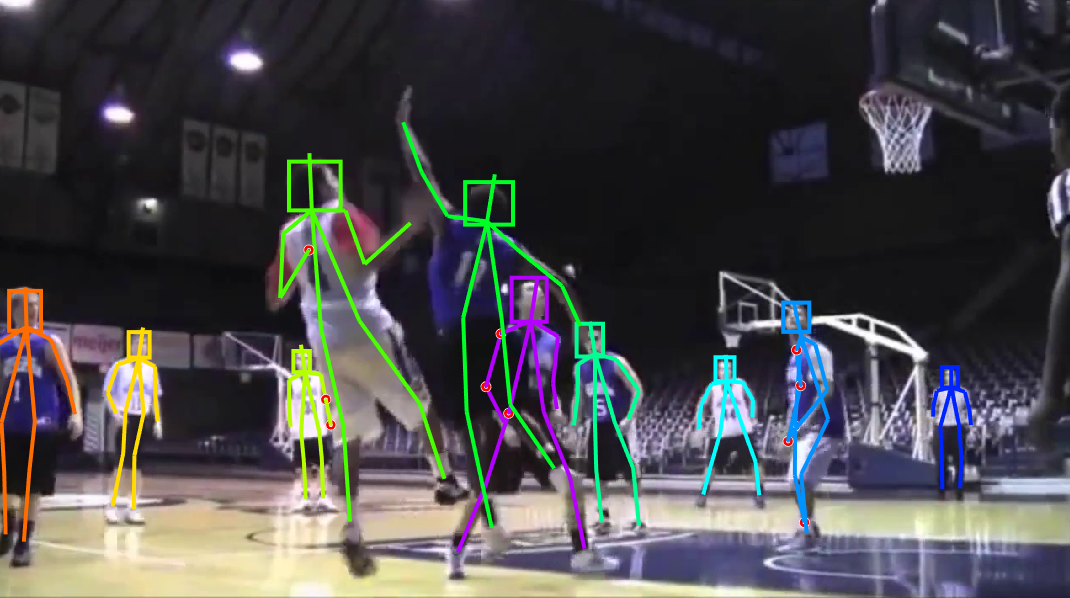} 
  \caption{Example frames and annotations from the proposed Multi-Person PoseTrack dataset.   \vspace{-5mm}}
    \label{fig:annotation_examples}
\end{figure}

Since there exists no dataset that provides annotations to quantitatively evaluate joint multi-person pose estimation and tracking, we also propose a new challenging \emph{Multi-Person PoseTrack} dataset as a second contribution of the paper. The dataset provides detailed and dense annotations for multiple persons in each video, as shown in Fig.~\ref{fig:annotation_examples}, and introduces new challenges to the field of pose estimation in videos. In order to evaluate the pose estimation and tracking accuracy, we introduce a new protocol that also deals with occluded body joints. We quantify the proposed method in detail on the proposed dataset, and also report results for several baseline methods.
The source code, pre-trained models and the dataset are publicly available.\footnote{\url{http://pages.iai.uni-bonn.de/iqbal_umar/PoseTrack/}}

%-------------------------------------------------------------------------
\section{Related Work}

Single person pose estimation in images has seen a remarkable progress over the past few years \cite{toshev2014deeppose, pishchulin2016deepcut, carreira2015human, wei2016convolutional, hu2016bottom, insafutdinov2016deepercut, newell2016eccv, bulat2016human, rafi2016bmvc}. However, all these approaches assume that only a single person is visible in the image, and cannot handle realistic cases where several people appear in the scene, and interact with each other. In contrast to single person pose estimation, multi-person pose estimation introduces significantly more challenges, since the number of persons in an image is not known a priori. Moreover, it is natural that persons occlude each other during interactions, and may also become partially truncated to various degrees. Multi-person pose estimation has therefore gained much attention recently \cite{eichner2010we, sun2011articulated, pishchulin2012articulated, yang_tpami2014, Ladicky_2013_CVPR, gkioxari2014using, chen2015parsing, belagiannis2015tpami, pishchulin2016deepcut, insafutdinov2016deepercut, Iqbal_ECCVw2016}. Earlier methods in this direction follow a two-staged approach \cite{pishchulin2012articulated, gkioxari2014using, chen2015parsing} by first detecting the persons in an image followed by a human pose estimation technique for each person individually. Such approaches are, however, applicable only if people appear well separated and do not occlude each other. Moreover, most single person pose estimation methods always output a fixed number of body joints and do not account for occlusion and truncation, which often is the case in multi-person scenarios. Other approaches address the problem using tree structured graphical models \cite{yang_tpami2014, sun2011articulated, eichner2010we, Ladicky_2013_CVPR}. However, such models struggle to cope with large pose variations, and are shown to be significantly outperformed by more recent methods based on Convolutional Neural Networks \cite{pishchulin2016deepcut, insafutdinov2016deepercut}. For example, \cite{pishchulin2016deepcut} jointly estimate the pose of all persons visible in an image, while also handling occlusion and truncation. The approach has been further improved by stronger part detectors and efficient approximations~\cite{insafutdinov2016deepercut}. The approach in \cite{Iqbal_ECCVw2016} also proposes a simplification of \cite{pishchulin2016deepcut} by tackling the problem locally for each person. However, it still relies on a separate person detector. 

Single person pose estimation in videos has also been studied extensively in the literature \cite{park_iccv2011, cherian_cvpr2014, zuffi_iccv2013, ramakrishna_cvpr2013, zuffi_iccv2013,  jain_accv2014, dong2015iccv, Pfister15a, georgia2016eccv, iqbal2017FG}. These approaches mainly aim to improve pose estimation by utilizing temporal smoothing constraints \cite{park_iccv2011, cherian_cvpr2014, dong2015iccv, ramakrishna_cvpr2013, georgia2016eccv} and/or optical flow information \cite{zuffi_iccv2013, jain_accv2014, Pfister15a}, but they are not directly applicable to videos with multiple potentially occluding persons. 

In this work we focus on the challenging problem of joint multi-person pose estimation and data association across frames. While the problem has not been studied quantitatively in the literature\footnote{Contemporaneously with this work, the problem has also been studied in \cite{Insafutdinov:2017:CVPR}}, there exist early works towards the problem \cite{Izadinia:2012:ECCV, Andriluka:2008:CVPR}. These approaches, however, do not reason jointly about pose estimation and tracking, but rather focus on multi-person tracking alone. The methods follow a multi-staged strategy, \ie they first estimate body part locations for each person separately and subsequently leverage body part tracklets to facilitate person tracking. We on the other hand propose to simultaneously estimate the pose of multiple persons and track them over time. To this end, we build upon the recent progress on multi-person pose estimation in images \cite{pishchulin2016deepcut, insafutdinov2016deepercut, Iqbal_ECCVw2016} and propose a joint objective for both problems. 

Previous datasets used to benchmark pose estimation algorithms in-the-wild are summarized in Tab.~\ref{tab:datasets}. While there exists a number of datasets to evaluate single person pose estimation methods in videos, such as \eg, \mbox{J-HMDB}~\cite{Jhuang_iccv2013} and Penn-Action~\cite{zhang2013actemes}, none of the video datasets provides annotations to benchmark multi-person pose estimation and tracking at the same time. To allow for a quantitative evaluation of this problem, we therefore also introduce a new ``Multi-Person PoseTrack'' dataset which provides pose annotations for multiple persons in each video to measure pose estimation accuracy, and also provides a unique ID for each of the annotated persons to benchmark multi-person pose tracking. The proposed dataset introduces new challenges to the field of human pose estimation and tracking since it contains a large amount of appearance and  pose variations, body part occlusion and truncation, large scale variations,  fast camera and person movements, motion blur, and a sufficiently large number of persons per video.   

\begin{figure*}[t]
  \centering
    \begin{tabular}{c}
		\includegraphics[trim={0 2mm 0 2mm}, clip, width=0.85\linewidth]{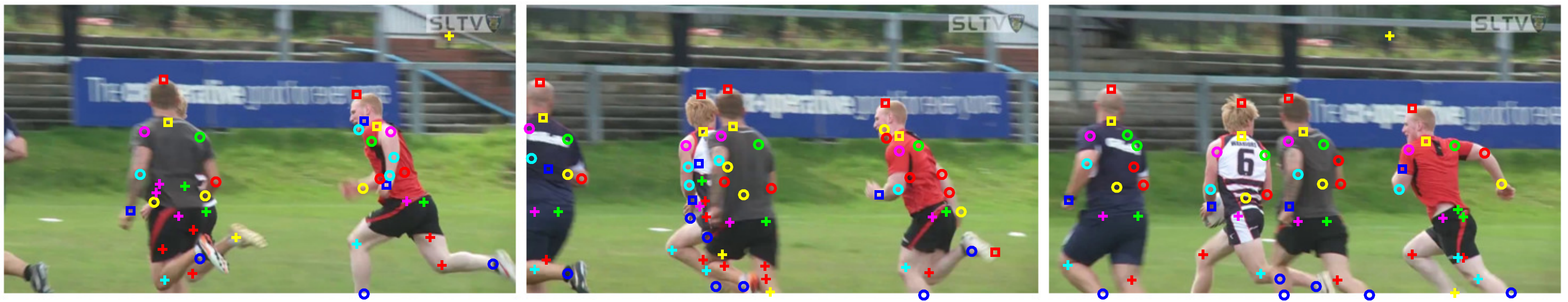} \\
  		\includegraphics[trim={0 2mm 0 2mm}, clip, width=0.85\linewidth]{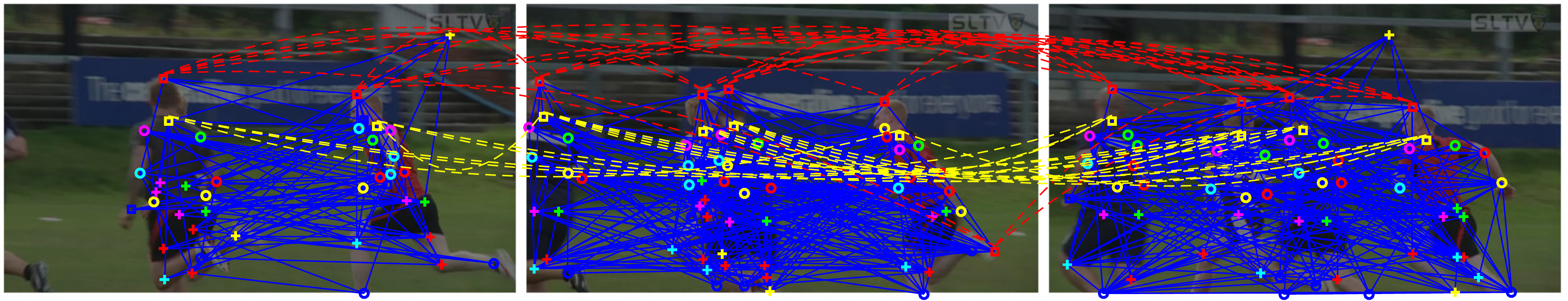} \\
	    \includegraphics[trim={0 2mm 0 2mm}, clip, width=0.85\linewidth]{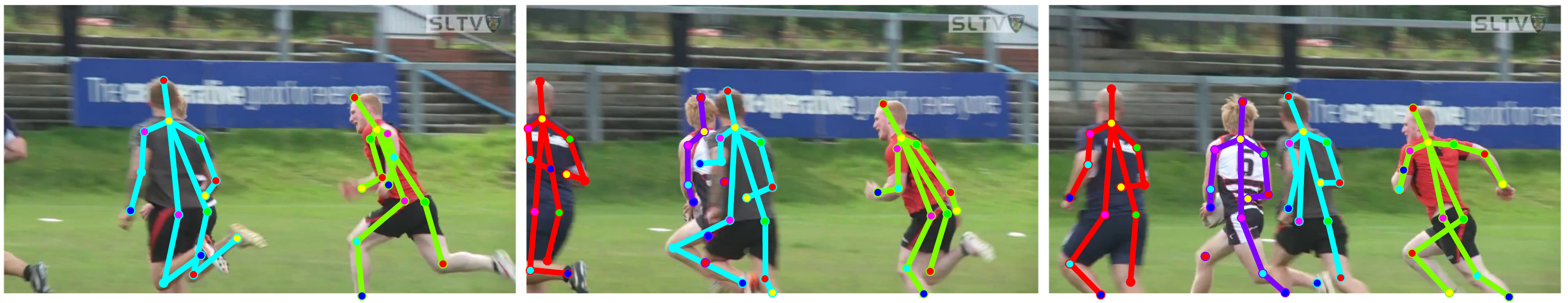} \\
    \end{tabular}
  \caption{{\bf Top:} Body joint detection hypotheses shown for three frames. {\bf Middle:} Spatio-temporal graph with spatial edges (blue) and temporal edges for head (red) and neck (yellow). We only show a subset of the edges. {\bf Bottom:} Estimated poses for all persons in the video. Each color corresponds to a unique person identity. 
  }
    \label{fig:overview}
    \vspace{-2mm}
\end{figure*}

\section{Multi-Person Pose Tracking}
\label{sec:joint_pose_estimation_and_tracking}

Our method jointly solves the problem of multi-person pose estimation and tracking for all persons appearing in a video together. We first generate a set of joint detection candidates in each video as illustrated in Fig.~\ref{fig:overview}. From the detections, we build a graph consisting of spatial edges connecting the detections within a frame and temporal edges connecting detections of the same joint type over frames.   
We solve the problem using integer linear programming (ILP) whose feasible solution provides the pose estimate for each person in all video frames, and also performs person association across frames.
We first introduce the proposed method and discuss the proposed dataset for evaluation in Sec.~\ref{sed:data}.

\subsection{Spatio-Temporal Graph}\label{sec:graph}
Given a video sequence $\mathcal{F}$ containing an arbitrary number of persons, we generate a set of body joint detection candidates $D = \{D_f\}_{f \in \mathcal{F}} $ where $D_f$ is the set for frame~$f$. Every detection $d\in D$ at location $\bold{x}^f_{d} \in \mathbb{R}^2$ in frame $f$ belongs to a joint type $j \in \mathcal{J} = \{1, \dots, J\}$. Additional details regarding the used detector will be provided in Sec.~\ref{sec:optimization_costs}.

For multi-person pose tracking, we aim to identify the joint hypotheses that belong to an individual person in the entire video. This can be formulated by a graph structure $G = (D,E)$ where $D$ is the set of nodes. The set of edges $E$ consists of two types of edges, namely spatial edges $E_s$ and temporal edges $E_t$. The spatial edges correspond to the union of edges of a fully connected graph for each frame, \ie 
\begin{equation}E_s = \bigcup_{f \in \mathcal{F}} E_s^f \;\text{and}\; E_s^f = \{ (d,d') : d{\neq}d' \land\ d, d' \in D_f\}.  
\end{equation}
Note that these edges connect joint candidates independently of the associated joint type $j$. The temporal edges connect only joint hypotheses of the same joint type over two different frames, \ie 
\begin{align}\nonumber E_t = \{ (d,d') :\ &j{=}j' \land\ d \in D_f \land\ d' \in D_{f'} \\
&\land\ 1{\leq}\vert f-f'\vert{\leq}\tau \land\ f, f' \in \mathcal{F}\}.
\end{align}
The temporal connections are not only modeled for neighboring frames, \ie $\vert f-f'\vert=1$, but we also take temporal relations up to $\tau$ frames into account to handle short-term occlusion and missing detections. The graph structure is illustrated in Fig.~\ref{fig:overview}.

\subsection{Graph Partitioning}
\label{sec:graph-partitioning}
By removing edges and nodes from the graph \mbox{$G=(D,E)$}, we obtain several partitions of the spatio-temporal graph and each partition corresponds to a tracked pose of an individual person. In order to solve the graph partitioning problem, we introduce the three binary vectors $v \in \{0,1\}^{\vert D \vert}$,  $s \in \{0,1\}^{\vert E_s \vert}$, and $t \in \{0,1\}^{\vert E_t \vert}$. Each binary variable implies if a node or edge is removed, \ie $v_d{=}0$ implies that the joint detection $d$ is removed. Similarly, $s_{(d_f,{d'}_f)}{=}0$ with $(d_f,{d'}_f)\in E_s$ implies that the spatial edge between the joint hypothesis $d$ and $d'$ in frame $f$ is removed while   
$t_{(d_f,{d'}_{f'})}{=}0$ with $(d_f,{d'}_{f'})\in E_t$ implies that the temporal edge between the joint hypothesis $d$ in frame $f$ and $d'$ in frame $f'$ is removed.

A partitioning is obtained by minimizing the cost function  
\begin{align}
\argmin_{v,s,t} &\left(\left\langle v, \phi \right\rangle + \left\langle s, \psi_s \right\rangle + \left\langle t, \psi_t \right\rangle\right) \label{eq:ilp_obj}\\
\left\langle v, \phi \right\rangle &= \sum_{d \in D} v_{d} \phi(d)  \\
\left\langle s, \psi_s \right\rangle &= \sum_{(d_f,{d'}_f) \in E_s} s_{(d_f,{d'}_f)} \psi_s(d_f,{d'}_f) \\
\left\langle t, \psi_t \right\rangle &= \sum_{(d_f,{d'}_{f'}) \in E_t} t_{(d_f,{d'}_{f'})} \psi_t(d_f,{d'}_{f'}) .
\end{align}
This means that we search for a graph partitioning such that the cost of the remaining nodes and edges is minimal. The cost for a node $d$ is defined by the unary term:  
\begin{equation}
\phi(d) = \log \dfrac{1-p_d}{p_d} \label{eq:unary} 
\end{equation}
where $p_d \in (0,1)$ corresponds to the probability of the joint hypothesis $d$. Note that $\phi(d)$ is negative when $p_d{>}0.5$ and detections with a high confidence are preferred since they reduce the cost function \eqref{eq:ilp_obj}. The cost for a spatial or temporal edge is defined similarly by
\begin{align}
\psi_s(d_f,{d'}_f) &= \log \dfrac{1-p^s_{(d_f,{d'}_f)}}{p^s_{(d_f,{d'}_f)}} \label{eq:sbinary} \\
\psi_t(d_f,{d'}_{f'}) &= \log \dfrac{1-p^t_{(d_f,{d'}_{f'})}}{p^t_{(d_f,{d'}_{f'})}} \label{eq:tbinary}.
\end{align}
While $p^s$ denotes the probability that two joint detections $d$ and $d'$ in a frame $f$ belong to the same person, $p^t$ denotes the probability that two detections of a joint in frame $f$ and $f'$ are the same. In Sec.~\ref{sec:optimization_costs} we will discuss how the probabilities $p_d$, $p^s_{(d_f,{d'}_f)}$, and $p^t_{(d_f,{d'}_{f'})}$ are learned.

In order to ensure that the feasible solutions of the objective \eqref{eq:ilp_obj} result in well defined body poses and valid pose tracks, we have to add additional constraints. The first set of constraints ensures that two joint hypotheses are associated to the same person ($s_{(d_f,{d'}_f)}{=}1$) only if both detections are considered as valid, \ie, $v_{{d}_f}{=}1$ and $v_{{d'}_f}{=}1$:  
\begin{equation}
s_{(d_f,{d'}_f)} \leq v_{d_f} \land\ s_{(d_f,{d'}_f)} \leq v_{{d'}_f} \quad \forall (d_f,{d'}_f) \in E_s. \label{eq:cont:consistency}
\end{equation}
The same holds for the temporal edges:
\begin{equation}
t_{(d_f,{d'}_{f'})} \leq v_{d_f} \land\ t_{(d_f,{d'}_{f'})} \leq v_{{d'}_{f'}} \quad \forall (d_f,{d'}_{f'}) \in E_t.
\end{equation}

\begin{figure}[t]
\captionsetup[subfigure]{justification=centering}
\centering
\begin{subfigure}[t]{0.11\textwidth}
\includegraphics[scale=0.65, center]{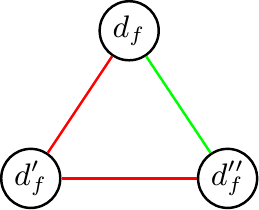}
\caption{}
\label{fig:constraints-illua}
\end{subfigure}
\hfill
\begin{subfigure}[t]{0.11\textwidth}
\includegraphics[scale=0.65,center]{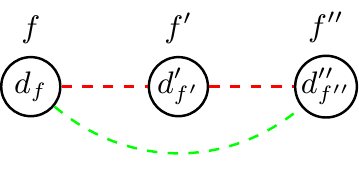}
\caption{}
\label{fig:constraints-illub}
\end{subfigure}
\hfill
\begin{subfigure}[t]{0.11\textwidth}
\includegraphics[scale=0.65,center]{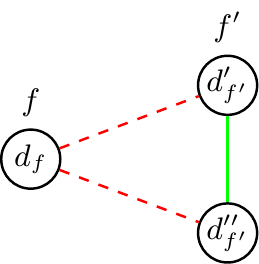}
\caption{}
\label{fig:constraints-illuc}
\end{subfigure}
\hfill
\begin{subfigure}[t]{0.11\textwidth}
\includegraphics[scale=0.65,center]{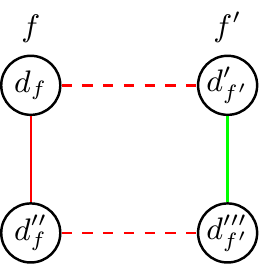}
\caption{}
\label{fig:constraints-illud}
\end{subfigure}
\caption{
\emph{(a)} The spatial transitivity constraints~\eqref{eq:cont:spatial-transitivity} ensure that if the two joint hypotheses $d_f$ and $d{''}_f$ are spatially connected to ${d'}_f$ (red edges) then the cost of the spatial edge between $d_f$ and $d{''}_f$ (green edge) also has to be added. \emph{(b)} The temporal transitivity constraints~\eqref{eq:cont:temporal-transitivity} ensure transitivity for temporal edges (dashed). \emph{(c)} The spatio-temporal transitivity constraints~\eqref{eq:cont:spatio-temporal-transitivity} model transitivity for two temporal edges and one spatial edge. \emph{(d)} The spatio-temporal consistency constraints~\eqref{eq:cont:spatio-temporal-consistency} ensure that if two pairs of joint hypotheses $(d_f,{d'}_{f'})$ and $(d{''}_{f},{d'''}_{f'})$ are temporally connected (dashed red edges) and $d_f$ and ${d''}_f$ are spatially connected (solid red edge) then the cost of the spatial edge between ${d'}_{f'}$ and $d{'''}_{f'}$ (solid green edge) also has to be added.
}
\label{fig:contraints}
\end{figure}

The second set of constraints are transitivity constraints in the spatial domain. Such transitivity constraints have been proposed for multi-person pose estimation in images~\cite{pishchulin2016deepcut, insafutdinov2016deepercut, Iqbal_ECCVw2016}. They enforce for any triplet of joint detection candidates $({d}_f,{d'}_f,{d''}_f)$ that if $d_f$ and ${d'}_f$ are associated to one person and ${d'}_f$ and ${d''}_f$ are also associated to one person, \ie $s_{(d_f,{d'}_f)}{=}1$ and $s_{({d'}_{f},{d''}_f)}{=}1$, then the edge $(d_f,{d''}_{f})$ should also be added:  
\begin{align}
s_{(d_f,{d'}_f)} + s_{({d'}_{f},{d''}_f)} - 1 &\leq s_{(d_f,{d''}_{f})} \label{eq:cont:spatial-transitivity}\\
\nonumber\quad\forall (d_f,{d'}_{f}), ({d'}_{f},{d''}_{f}) &\in E_s . 
\end{align}
An example of a transitivity constraint is illustrated in Fig.~\ref{fig:constraints-illua}. The transitivity constraints can be used to enforce that a human can have only one joint type $j$, \eg only one head. Let $d_f$ and ${d''}_f$ have the same joint type $j$ while ${d'}_f$ belongs to another joint type $j'$. Without transitivity constraints connecting $d_f$ and ${d''}_f$ with ${d'}_f$ 
might result in a low cost. The transitivity constraints, however, enforce that the binary cost $\psi_s(d_f,{d''}_{f})$ is added. 
To prevent poses with multiple joints, we thus only have to ensure that the binary cost $\psi_s(d,d'')$ is very high if $j{=}j''$. We discuss this more in detail in Sec.~\ref{sec:optimization_costs}. 

In contrast to previous work, we also have to ensure spatio-temporal consistency. Similar to the spatial transitivity constraints~\eqref{eq:cont:spatial-transitivity}, we can define temporal transitivity constraints:    
\begin{align}
t_{(d_f,{d'}_{f'})} + t_{({d'}_{f'},{d''}_{f''})} - 1 &\leq t_{(d_f,{d''}_{f''})}\label{eq:cont:temporal-transitivity} \\
\nonumber\forall (d_f,{d'}_{f'}), ({d'}_{f'},{d''}_{f''}) &\in E_t   . 
\end{align}

The last set of constraints are spatio-temporal constraints that ensure that the pose is consistent over time. We define two types of spatio-temporal constraints. The first type consists of a triplet of joint detection candidates $(d_f, {d'}_{f'}, d''_{f'})$ from two different frames $f$ and $f'$. The constraints are defined as,
\begin{align}
\nonumber t_{(d_f,{d'}_{f'})} + t_{({d}_{f},{d''}_{f'})} - 1 &\leq s_{({d'}_{f'},{d''}_{f'})} \\
t_{(d_f,{d'}_{f'})} + s_{({d'}_{f'},{d''}_{f'})} - 1 &\leq t_{({d}_{f},{d''}_{f'})} \label{eq:cont:spatio-temporal-transitivity} \\
\nonumber \forall ({d}_{f},{d'}_{f'}),  ({d}_{f},{d''}_{f'}) &\in E_t,  
%t_{(d_f,{d'}_{f'})} + s_{({d}_{f},{d''}_{f})} - 1 &\leq t_{({d''}_{f},{d''}_{f'})}\label{eq:cont:spatio-temporal-transitivity} \\
%\forall ({d}_{f},{d'}_{f'})\in E_t, \forall ({d}_{f},{d''}_{f}) &\in E_s,
\end{align}
and enforce transitivity for two temporal edges and one spatial edge. 
The second type of spatio-temporal constraints are based on quadruples of joint detection candidates $({d}_f,{d'}_{f'},{d''}_f,{d'''}_{f'})$ from two different frames $f$ and $f'$. The spatio-temporal constraints ensure that if $({d}_f,{d'}_{f'})$ and $({d''}_{f},{d'''}_{f'})$ are temporally connected and $({d}_f,{d''}_{f})$ are spatially connected then the spatial edge $({d'}_{f'},{d'''}_{f'})$ has to be added:  
\begin{align}
\nonumber t_{(d_f,{d'}_{f'})} + t_{({d''}_{f},{d'''}_{f'})} + s_{(d_f,{d''}_f)} - 2 &\leq s_{({d'}_{f'},{d'''}_{f'})} \\
t_{(d_f,{d'}_{f'})} + t_{({d''}_{f},{d'''}_{f'})} + s_{({d'}_{f'},{d'''}_{f'})} - 2 &\leq s_{({d}_{f},{d''}_{f})}  \label{eq:cont:spatio-temporal-consistency} \\
\nonumber\forall (d_f,{d'}_{f'}), ({d''}_{f},{d'''}_{f'}) &\in E_t  . 
\end{align}
An example of both types of spatio-temporal constraint can be seen in Fig.~\ref{fig:constraints-illuc} and Fig.~\ref{fig:constraints-illud}, respectively.

\subsection{Optimization}
We optimize the objective \eqref{eq:ilp_obj} with the branch-and-cut algorithm of the ILP solver Gurobi. To reduce the runtime for long sequences, we process the video batch-wise where each batch consists of $k=31$ frames. For the first $k$ frames, we build the spatio-temporal graph as discussed and optimize the objective \eqref{eq:ilp_obj}. We then continue to build a graph for the next $k$ frames and add the previously selected nodes and edges to the graph, but fix them such that they cannot be removed anymore. Since the graph partitioning produces also small partitions, which usually correspond to clusters of false positive joint detections, we remove any partition that is shorter than 7 frames or has less than 6 nodes per frame on average. 

\vspace{-1mm}
\subsection{Potentials}
\label{sec:optimization_costs}
In order to compute the unaries $\phi$ \eqref{eq:unary} and binaries $\psi$ \eqref{eq:sbinary},\eqref{eq:tbinary}, we have to learn the probabilities $p_d$, $p^s_{(d_f,{d'}_f)}$, and $p^t_{(d_f,{d'}_{f'})}$. 

The probability $p_d$ is given by the confidence of the joint detector. As joint detector, we use the publicly available pre-trained CNN~\cite{insafutdinov2016deepercut} trained on the MPII Multi-Person Pose dataset \cite{pishchulin2016deepcut}. In contrast to~\cite{insafutdinov2016deepercut}, we do not assume that any scale information is given. We therefore apply the detector to an image pyramid with 4 scales $\gamma \in \{0.6, 0.9, 1.2, 1.5\}$. For each detection $d$ located at $\bold{x}^f_d$, we compute a quadratic bounding box $B_d = \{\bold{x}^f_d, h_d\}$. We use $h_d=\frac{70}{\gamma}$ for the width and height. To reduce the number of detections, we remove all bounding boxes that have an intersection-over-union (IoU) ratio over 0.7 with another bounding box that has a higher detection confidence.   

The spatial probability $p^s_{(d_f,{d'}_f)}$ depends on the joint types $j$ and $j'$ of the detections. If $j{=}j'$, we define \mbox{$p^s_{(d_f,{d'}_f)}{=}\text{IoU}(B_d, B_{d'})$}. This means that a joint type $j$ cannot be added multiple times to a person except if the detections are very close. If a partition includes detections of the same type in a single frame, the detections are merged by computing the weighted mean of the detections, where the weights are proportional to $p_d$. If $j{\neq}j'$, we use the pre-trained binaries~\cite{insafutdinov2016deepercut} after a scale normalization.           

The temporal probability $p^t_{(d_f,{d'}_{f'})}$ should be high if two detections of the same joint type at different frames belong to the same person. To that end, we build on the idea recently used in multi-person tracking \cite{tang-2016} and compute dense correspondences between two frames using DeepMatching \cite{weinzaepfel:hal-00873592}. Let $K_{d_f}$ and $K_{{d'}_{f'}}$ be the sets of matched key-points inside the bounding boxes $B_{d_f}$ and $B_{{d'}_{f'}}$ and $\KU_{{d}{{d'}}}{=}\vert K_{d_f} \cup K_{{d'}_{f'}} \vert$ and $\KI_{{d}{{d'}}}{=}\vert K_{d_f} \cap K_{{d'}_{f'}} \vert$ the union and intersection of these two sets. We then form a feature vector by $\{ \KI/\KU, \min(p_d, p_{d'}), \Delta\bold{x}_{dd'}, \Vert \Delta\bold{x}_{dd'} \Vert \}$ where \mbox{$\Delta\bold{x}_{dd'} = \bold{x}^f_d - \bold{x}^{f'}_{d'}$}. We also append the feature vector with non-linear terms as done in \cite{tang-2016}. The mapping from the feature vector to the probability $p^t_{(d_f,{d'}_{f'})}$ is obtained by logistic regression.

\newcommand{\squeezeCol}{\vspace{-0.1cm}}

\begin{table}
\centering
\scriptsize
\begin{tabularx}{\columnwidth}{Xssssu}
\toprule
Dataset & \rotatebox[origin=c]{90}{\parbox{1.5cm}{\centering \squeezeCol Video-labeled \\ poses \squeezeCol}} & \rotatebox[origin=c]{90}{multi-person} & \rotatebox[origin=c]{90}{\parbox{1.5cm}{\centering \squeezeCol Large scale \\ variation \squeezeCol}} & \rotatebox[origin=c]{90}{\parbox{1.5cm}{\centering \squeezeCol variable \\ skeleton size \squeezeCol}} & \rotatebox[origin=c]{90}{\# of persons}\\
\midrule
Leeds Sports \cite{Ever10} & &  & & & 2000 \\
MPII Pose \cite{andriluka_cvpr2014} &   &  & \checkmark & \checkmark & 40,522\\
We Are Family \cite{eichner2010we} &   & \checkmark &  &  & 3131\\
MPII Multi-Person Pose  \cite{pishchulin2016deepcut} &   & \checkmark &  \checkmark & \checkmark & 14,161\\
MS-COCO Keypoints \cite{lin2014microsoft} &  & \checkmark & \checkmark & \checkmark & 105,698 \\
\midrule
J-HMDB \cite{Jhuang_iccv2013} & \checkmark  & & \checkmark & \checkmark & 32,173\\
Penn-Action \cite{zhang2013actemes} & \checkmark & & \checkmark & & 159,633\\ 
VideoPose \cite{sap_cvpr2011} & \checkmark  & & & & 1286\\ 
Poses-in-the-wild \cite{cherian_cvpr2014}  & \checkmark & & & & 831\\ 
YouTube Pose \cite{charles2016cvpr} & \checkmark  & &  & & 5000\\ 
FYDP \cite{shen2014_eccv2014} & \checkmark  & & & & 1680\\ 
UYDP \cite{shen2014_eccv2014} & \checkmark  & & & & 2000\\ 
\midrule
\textbf{Multi-Person PoseTrack} & \checkmark & \checkmark & \checkmark & \checkmark & 16,219\\ 
\bottomrule
\end{tabularx}
\caption{A comparison of PoseTrack dataset with the existing related datasets for human pose estimation in images and videos.\vspace{-3mm}}
\label{tab:datasets}
\end{table}

\section{The Multi-Person PoseTrack Dataset}\label{sed:data}
In this section we introduce our new dataset for multi-person pose estimation in videos. The MPII Multi-Person Pose \cite{andriluka_cvpr2014} is currently one of the most popular benchmarks for multi-person pose estimation in images, and covers a wide range of activities. For each annotated image, the dataset also provides unlabeled video clips ranging 20 frames both forward and backward in time relative to that image. For our video dataset, we manually select a subset of all available videos that contain multiple persons and cover a wide variety of person-person or person-object interactions. Moreover, the selected videos are chosen to contain a large amount of body pose appearance and scale variation, as well as body part occlusion and truncation. The videos also contain severe body motion, \ie, people occlude each other, re-appear after complete occlusion, vary in scale across the video, and also significantly change their body pose. The number of visible persons and body parts may also vary during the video. 
The duration of all provided video clips is exactly 41 frames. To include longer and variable-length sequences, we downloaded the original raw video clips using the provided URLs and obtained an additional set of videos. To prevent an overlap with the existing data, we only considered sequences that are at least 150 frames apart from the training samples, and followed the same rationale as above to ensure diversity.

In total, we compiled a set of 60 videos with the number of frames per video ranging between 41 and 151. The number of persons ranges between 2 and 16 with an average of more than 5 persons per video sequence, totaling over 16,000 annotated poses. The person heights are between 100 and 1200 pixels. We split the dataset into a training and testing set with an equal number of videos.

\subsection{Annotation}
As in~\cite{andriluka_cvpr2014}, we annotate 14 body joints and a rectangle enclosing the person's head. The latter is required to estimate the absolute scale which is used for evaluation. We assign a unique identity to every person appearing in the video. This person ID remains the same throughout the video until the person moves out of the field-of-view. Since we do not target person re-identification in this work, we assign a new ID if a person re-appears in the video.  We also provide occlusion flags for all body joints. A joint is marked occluded if it was in the field-of-view but became invisible due to an occlusion. Truncated joints, \ie those outside the image border limits, are not annotated, therefore, the number of joints per person varies across the dataset. Very small persons were zoomed in to a reasonable size to accurately perform the annotation. To ensure a high quality of the annotation, all annotations were performed by trained in-house workers, following a clearly defined protocol. An example annotation can be seen in Fig. \ref{fig:annotation_examples}.

\subsection{Experimental setup and evaluation metrics}
\label{sec:metrics}
Since the problem of simultaneous multi-person pose estimation and person tracking has not been quantitatively evaluated in the literature, we define a new evaluation protocol for this problem. To this end, we follow the best practices followed in both multi-person pose estimation \cite{pishchulin2016deepcut} and multi-target tracking \cite{milan2016mot16}. In order to evaluate whether a part is predicted correctly, we use the widely adopted PCKh (head-normalized probability of correct keypoint) metric \cite{andriluka_cvpr2014}, which considers a body joint to be correctly localized if the predicted location of the joint is within a certain threshold from the true location. Due to the large scale variation of people across videos and even within a frame, this threshold needs to be selected adaptively, based on the person's size. To that end,~\cite{andriluka_cvpr2014} propose to use 30\% of the head box diagonal.
We have found this threshold to be too relaxed because recent pose estimation approaches are capable of predicting the joint locations rather accurately. Therefore, we use a more strict evaluation with a 20\% threshold.

Given the joint localization threshold for each person, we compute two sets of evaluation metrics, one adopted from the multi-target tracking literature~\cite{Yang:2012:CVPR, Choi:2015:ICCV, milan2016mot16} to evaluate multi-person pose tracking, and one which is commonly used for evaluating multi-person pose estimation~\cite{pishchulin2016deepcut}.

\noindent\textbf{Tracking.}
To evaluate multi-person pose tracking, we consider each joint trajectory as one individual target,\footnote{Note that only joints of the same type are matched.} and compute multiple measures.
First, the CLEAR MOT metrics~\cite{Bernardin:2008:CLE} provide the tracking accuracy (MOTA) and tracking precision (MOTP). The former is derived from three types of error ratios: false positives, missed targets, and identity switches (IDs). These are linearly combined to produce a normalized accuracy where 100\% corresponds to zero errors. MOTP measures how precise each object, or in our case each body joint, has been localized \wrt the ground-truth.
Second, we report trajectory-based measures proposed in~\cite{Li:2009:CVPR}, that count the number of mostly tracked (MT) and mostly lost (ML) tracks. A track is considered mostly tracked if it has been recovered in at least 80\% of its length, and mostly lost if more than 80\% are not tracked.
For completeness, we also compute the number of times a ground-truth trajectory is fragmented (FM). 

\noindent\textbf{Pose.}
For measuring frame-wise multi-person pose accuracy, we use \emph{Mean Average Precision} (mAP) as is done in \cite{pishchulin2016deepcut}.
The protocol to evaluate multi-person pose estimation in \cite{pishchulin2016deepcut} assumes that the rough scale and location of a group of persons is known during testing \cite{pishchulin2016deepcut}, which is not the case in realistic scenarios, and in particular in videos. We therefore propose to make no assumption during testing and evaluate the predictions without rescaling or shifting them according to the ground-truth.

\noindent\textbf{Occlusion handling.}
Both of the aforementioned protocols to measure pose estimation and tracking accuracy do not consider occlusion during evaluation, and penalize if an occluded target that is annotated in the ground-truth is not correctly estimated \cite{milan2016mot16, pishchulin2016deepcut}. This, however, discourages methods that either detect occlusion and do not predict the occluded joints or approaches that predict the joint position even for occluded joints. We want to provide a fair comparison for both types of occlusion handling. We therefore extend both measures to incorporate occlusion information explicitly. To this end, we first assign each person to one of the ground-truth poses based on the PCKh measure as done in \cite{pishchulin2016deepcut}. For each matched person, we consider an occluded joint correctly estimated 
either if \emph{a)} it is predicted at the correct location despite being occluded,
or \emph{b)} it is not predicted at all. Otherwise, the prediction is considered as a false positive.

\section{Experiments}
In this section we evaluate the proposed method for joint multi-person pose estimation and tracking on the newly introduced Multi-Person PoseTrack dataset. 

\begin{table}
\centering
\scriptsize
\begin{tabularx}{\columnwidth}{Xssssssstt}
\toprule
%Exp10: Head Only
\textbf{Method} & \textbf{Rcll}  &  \textbf{Prcn} &  \textbf{MT}  &  \textbf{ML} &  \textbf{IDs}  &  \hspace*{-1mm} \textbf{FM}   & \hspace*{-3mm} \textbf{ MOTA} &   \hspace*{-1mm} \textbf{MOTP}   \\
 &  ~~$\uparrow$ &  ~~$\uparrow$ &   ~$\uparrow$  &  ~$\downarrow$ &  ~$\downarrow$  &    ~~$\downarrow$  &   $\uparrow$ &  $\uparrow$  \\
\midrule
\multicolumn{9}{c}{\textit{Impact of temporal connection density}} \\
\midrule
HT & 57.6 &  66.0 & 632 & 623 &  674 & 5080 &  27.2 &   56.1 \\ 
%Exp11:  head, neck, shoulders
HT:N:S & 62.7 &  64.9 & 760 & 510 &  470 & 5557 &  28.2 &   55.8    \\ 
%Exp12:  head, neck, shoulders
HT:N:S:H & 63.1 &  64.5 & 774 & 494 &  478 & 5564 &  27.8 &   55.7 \\ 
%Exp13: head, wrists, ankles
HT:W:A & 62.8 &  64.9  & 758 & 526 &  516 & 5458 &  28.2 &   55.8 \\   
\midrule
\multicolumn{9}{c}{\textit{Impact of the length of temporal connection ($\tau$)}} \\  
\midrule
HT:N:S ($\tau=1$) & 62.7 &  64.9 & 760 & 510 &  470 & 5557 &  28.2 &   55.8  \\ 
HT:N:S ($\tau=3$) & 63.0 &  64.8 & 775 & 502 &  431 & 5629 &  28.2 &   55.7  \\ 
HT:N:S ($\tau=5$) & 62.8 &  64.7 & 763 & 508 &  381 & 5676 &  28.0 &   55.7  \\   
\midrule
\multicolumn{9}{c}{\textit{Impact of the constraints}} \\  
\midrule
All & 63.0 &  64.8 & 775 & 502 &  431 & 5629 &  28.2 &   55.7 \\ 
All $\setminus$ spat. transitivity & 22.2 &  76.0 & 115 & 1521 &   39 & 3947 &  15.1 &   58.0 \\ 
All $\setminus$ temp. transitivity & 60.3 &  65.1 & 712 & 544 &  268 & 5610 &  27.7 &   55.8 \\ 
All $\setminus$ spatio-temporal &  55.1 &  64.1 & 592 & 628 &  262 & 5444 &  23.9 &   55.7 \\  
\midrule
\multicolumn{9}{c}{\textit{Comparison with the Baselines}} \\  
\midrule
\bf Ours  & \textbf{63.0} &  \textbf{64.8}  & \textbf{775}  & \textbf{502} &  431 & 5629 &  \textbf{28.2} &   \textbf{55.7}  \\  
% Exp101: Detection + Tracking + CPM Pose Estimation + ILP
\multicolumn{9}{l}{BBox-Tracking \cite{tang-2016, ren2015faster}} \\  

\quad + LJPA \cite{Iqbal_ECCVw2016} & 58.8 &  64.8 & 716 & 646 &  \textbf{319} & \textbf{5026} &  26.6 &   53.5 \\ 
% Exp101: Detection + Tracking + CPM Pose Estimation
\quad + CPM \cite{wei2016convolutional} &  60.1 &  57.7 & 754 & 611 &  347 & 4969 &  15.6 &   53.4   \\ 
\bottomrule
\end{tabularx}
\caption{Quantitative evaluation of multi-person pose-tracking using common multi-object tracking metrics. Up and down arrows indicate whether higher or lower values for each metric are better. The first three blocks of the table present an ablative study on design choices \wrt joint selection, temporal edges, and constraints. The bottom part compares our final result with two strong baselines described in the text. HT:Head Top, N:Neck, S:Shoulders, W:Wrists, A:Ankles \vspace{-4mm}}
\label{tab:mot_results}
\end{table}

\subsection{Multi-Person Pose Tracking}
The results for multi-person pose tracking (MOT CLEAR metrics) are reported in \Tab~\ref{tab:mot_results}. To find the best setting, we first perform a series of experiments, investigating the influence of temporal connection density, temporal connection length, and inclusion of different constraint types.

We first examine the impact of different joint combinations for temporal connections.  Connecting only the Head Tops (HT) between frames results in a Multi-Object Tracking Accuracy (MOTA) of $27.2$ with a recall and precision of $57.6\%$ and $66.0\%$, respectively. Adding Neck and Shoulder (HT:N:S) detections for temporal connections improves the MOTA score to $28.2$, while also improving the recall from $57.6\%$ to $62.7\%$.  Adding more temporal connections also increases other metrics such as MT, ML, and also results in a lower number of ID switches (IDs) and fragments (FM). However, increasing the number of joints for temporal edges even further (HT:N:S:H) results in a slight decrease in performance. This is most likely due to the weaker DeepMatching correspondences between hip joints, which are difficult to match. When only the body extremities (HT:W:A) are used for temporal edges, we obtain a similar MOTA as for (HT:N:S), but slightly worse other tracking measures. Considering the MOTA performance and the complexity of our graph structure, we use (HT:N:S) as our default setting. 

Instead of considering only neighboring frames for temporal edges, we also evaluate the tracking performance while introducing longer-range temporal edges of up to $3$ and $5$ frames.
Adding temporal edges between detections that are at most three frames $(\tau\!=\!3)$ apart improves the performance only slightly, whereas increasing the distance even further $(\tau\!=\!5)$ worsens the performance. For the rest of our experiments we therefore set $\tau=3$. 

To evaluate the proposed optimization objective \eqref{eq:ilp_obj} for joint multi-person pose estimation and tracking in more detail, we have quantified the impact of various kinds of constraints \eqref{eq:cont:consistency}-\eqref{eq:cont:spatio-temporal-consistency} enforced during the optimization. To this end, we remove one type of constraints at a time and solve the optimization problem. As shown in Tab.~\ref{tab:mot_results}, all types of constraints are important to achieve best performance, with the spatial transitivity constraints playing the most crucial role. This is expected since these constraints ensure that we obtain valid poses without multiple joint types assigned to one person.    
Temporal transitivity and spatio-temporal constraints also turn out to be important to obtain good results. Removing either of the two significantly decreases the recall, resulting in a drop in MOTA. 

Since we are the first to report results on the Multi-Person PoseTrack dataset, we also develop two baseline methods by using the existing approaches. For this, we rely on a state-of-the-art method for multi-person pose estimation in images \cite{Iqbal_ECCVw2016}. The approach uses a person detector \cite{ren2015faster} to first obtain person bounding box hypotheses, and then estimates the pose for each person independently. We extend it to videos as follows. We first generate person bounding boxes for all frames in the video using a state-of-the-art person detector (Faster R-CNN \cite{ren2015faster}), and perform person tracking using a state-of-the-art person tracker \cite{tang-2016} and train it on the training set of the Multi-Person PoseTrack Dataset. We also discard all tracks that are shorter than 7 frames. The final pose estimates are obtained by using the Local Joint-to-Person Association (LJPA) approach proposed by \cite{Iqbal_ECCVw2016} for each person track. We also report results when Convolutional Pose Machines (CPM) \cite{wei2016convolutional} are used instead. 
Since CPM does not account for joint occlusion and truncation, the MOTA score is significantly lower than for LJPA. LJPA~\cite{Iqbal_ECCVw2016} improves the performance, but remains inferior \wrt most measures compared to our proposed method. In particular, our method achieves the highest MOTA and MOTP scores. The former is due to a significantly higher recall, while the latter is a result of a more precise part localization. Interestingly, the person bounding-box tracking based baselines achieve a lower number of ID switches. We believe that this is primarily due to the powerful multi-target tracking approach~\cite{tang-2016}, which can handle person identities more robustly.

\begin{table}

\scriptsize
\begin{tabularx}{\columnwidth}{lssssssssss}
\toprule
\textbf{Method} & \textbf{Head} & \textbf{Sho} & \textbf{Elb} & \textbf{Wri} & \textbf{Hip} & \textbf{Knee}  &  \textbf{Ank} & \textbf{mAP}\\ \midrule
\multicolumn{9}{c}{\textit{Impact of the temporal connection density}} \\ 
\midrule
%Exp10: Head Only. Frames = 1
HT & 52.5  & 47.0  & 37.6  & 28.2  & 19.7  & 27.8 & 27.4 & 34.3 \\ 
%Exp11:  head, neck, shoulders. Frames = 1\n
HT:N:S & 56.1  & 51.3  & 42.1  & 31.2  & 22.0  & 31.6 & 31.3 & 37.9 \\ 
%Exp12: head, neck, shoulders, hips. Frames = 1
HT:N:S:H & 56.3  & 51.5  & 42.2  & 31.4  & 21.7  & 31.6 & 32.0 & 38.1 \\ 
%Exp13: head, wrists, ankles. Frames = 1
HT:W:A & 56.0  & 51.2  & 42.2  & 31.6  & 21.6  & 31.2 & 31.7 & 37.9 \\ 
\midrule
\multicolumn{9}{c}{\textit{Impact of the length of temporal connection ($\tau$)}} \\ 
\midrule
%Exp3:  head, neck, shoulders, Frames = 3\n
HT:N:S $(\tau=1)$ & 56.1  & 51.3  & 42.1  & 31.2  & 22.0  & 31.6 & 31.3 & 37.9 \\ 
HT:N:S $(\tau=3)$ & 56.5  & 51.6  & 42.3  & 31.4  & 22.0  & 31.9 & 31.6 & 38.2 \\ 
%Exp7:  head, neck, shoulders. Frames = 5
HT:N:S $(\tau=5)$ & 56.2  & 51.3  & 41.8  & 31.1  & 22.0  & 31.4 & 31.5 & 37.9 \\  
\midrule
\multicolumn{9}{c}{\textit{Impact of the constraints}} \\ 
\midrule 
%Exp3:  head, neck, shoulders, Frames = 3
All & 56.5  & 51.6  & 42.3  & 31.4  & 22.0  & 31.9 & 31.6 & 38.2 \\ 
%Exp14:  head, neck, shoulders, Frames = 3. Without Spatial Transitivity Constraints
All $\setminus$ spat. transitivity & 7.8  & 10.1  & 7.2  & 4.6  & 2.7  & 4.9 & 5.9 & 6.2 \\ 
%Exp15:  head, neck, shoulders, Frames = 3. Without Temporal Transitivity Constraints
All $\setminus$ temp. transitivity & 50.5  & 46.8  & 37.5  & 27.6  & 20.3  & 30.1 & 28.7 & 34.5 \\ 
%Exp16:  head, neck, shoulders, Frames = 3. Without Spatio-Temporal Constraints
All $\setminus$ spatio-temporal & 42.3  & 40.8  & 32.8  & 24.3  & 17.0  & 25.3 & 22.4 & 29.3 \\ 
\midrule
\multicolumn{9}{c}{\textit{Comparison with the state-of-the-art}} \\  
\midrule
%Exp3:  head, neck, shoulders, Frames = 3\n
Ours & \textbf{56.5}  & 51.6  & \textbf{42.3}  & \textbf{31.4}  & 22.0  & \textbf{31.9} & \textbf{31.6} & \textbf{38.2} \\ 
%Exp101: Detection + Tracking + CPM Pose Estimation + ILP
\multicolumn{9}{l}{BBox-Detection \cite{ren2015faster}} \\  
\quad~ + LJPA \cite{Iqbal_ECCVw2016} & 50.5  & 49.3  & 38.3  & 33.0  & 21.7  & 29.6 & 29.2 & 35.9 \\ 
%Exp102: Detection + Tracking + CPM Pose Estimation&
\quad~ + CPM \cite{wei2016convolutional} & 48.8  & 47.5  & 35.8  & 29.2  & 20.7  & 27.1 & 22.4 & 33.1 \\ 
%DeeperCut
DeeperCut \cite{insafutdinov2016deepercut} & 56.2  & \textbf{52.4}  & 40.1  & 30.0  & \textbf{22.8}  & 30.5 & 30.8 & 37.5 \\ 
\bottomrule
\end{tabularx}
\caption{Quantitative evaluation of multi-person pose estimation (mAP). HT:Head Top, N:Neck, S:Shoulders, W:Wrists, A:Ankles \vspace{-8mm}}
\label{tab:map_results}
\end{table}

\begin{figure*}[t]
\captionsetup[subfigure]{justification=centering}
\centering
\begin{subfigure}[t]{0.33\textwidth}
\includegraphics[scale=0.73, center]{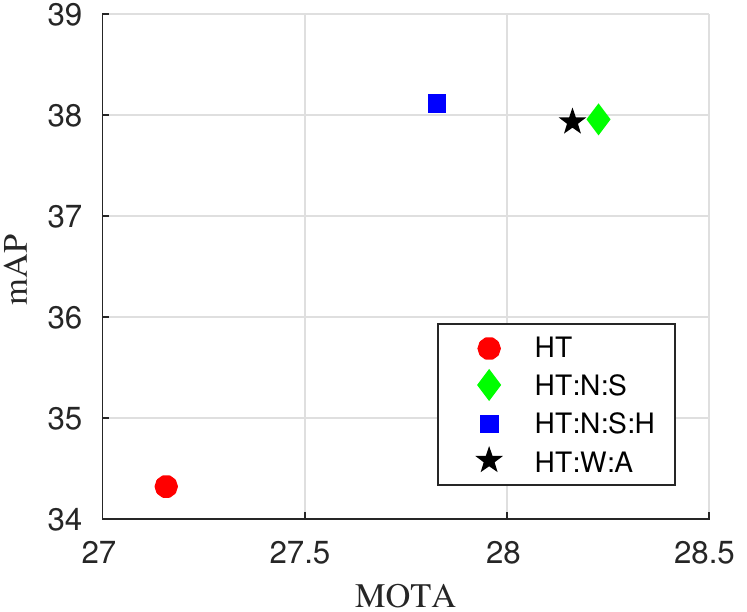}
\label{fig:contraints-a}
\end{subfigure}
\begin{subfigure}[t]{0.33\textwidth}
\includegraphics[scale=0.73,center]{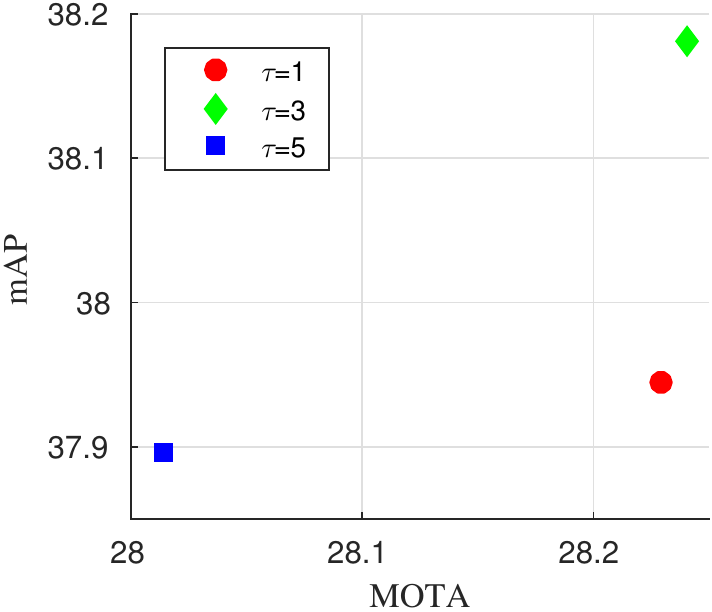}
\label{fig:contraints-b}
\end{subfigure}
\begin{subfigure}[t]{0.33\textwidth}
\includegraphics[scale=0.73,center]{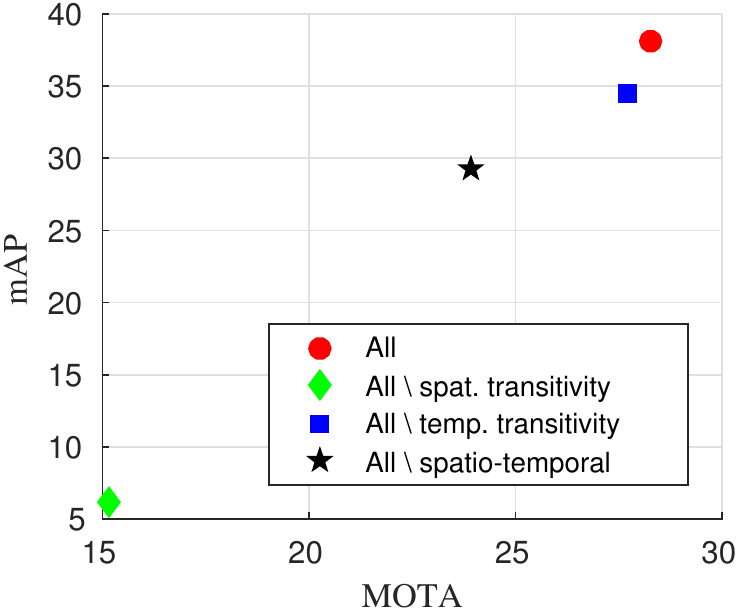}
\label{fig:contraints-b}
\end{subfigure}
\vspace{-4mm}
\caption{\textbf{Left} Impact of the the temporal edge density. \textbf{Middle} Impact of the length of temporal edges. \textbf{Right} Impact of different constraint types.\vspace{-5mm}}
\label{fig:MOTAvsMAP}
\end{figure*}

\subsection{Frame-wise Multi-Person Pose Estimation}

The results for frame-wise multi-person pose estimation (mAP) are summarized in Tab.~\ref{tab:map_results}. Similar to the evaluation for pose tracking, we evaluate the impact of spatio-temporal connection density, length of temporal connections and the influence of different constraint types. Having connections only between Head Top (HT) detections results in a mAP of $34.3\%$. As for pose tracking, introducing temporal connections for Neck and Shoulders (HT:N:S) results in a higher accuracy and improves the mAP from $34.3\%$ to $37.9\%$. The mAP elevates slightly more when we also incorporate connections for hip joints (HT:N:S:H). This is in contrast to pose tracking where MOTA dropped slightly when we also use connections for hip joints. As before, inclusion of edges between all detections that are in the range of $3$ frames improves the performance, while increasing the distance further $(\tau=5)$ starts to deteriorate the performance. A similar trend can also been seen for the impact of different types of constraints. The removal of spatial transitivity constraints results in a drastic decrease in pose estimation accuracy. Without temporal transitivity constraints or spatio-temporal constraints the pose estimation accuracy drops  by more than $3\%$ and $8\%$, respectively. This once again indicates that all types of constraints are essential to obtain better pose estimation and tracking performance. 

We also compare the proposed method with the state-of-the-art approaches for multi-person pose estimation in images. Similar to \cite{Iqbal_ECCVw2016}, we use Faster R-CNN \cite{ren2015faster} as person detector, and use the provided codes for LJPA \cite{Iqbal_ECCVw2016} and CPM \cite{wei2016convolutional} to process each bounding box detection independently. We can see that person bounding box based approaches significantly underperform as compared to the proposed method. We also compare with the state-of-the-art method DeeperCut \cite{insafutdinov2016deepercut}. The approach, however, requires the rough scale of the persons during testing. For this, we use the person detections obtained from \cite{ren2015faster} to compute the scale using the median scale of all detected persons.

Our approach achieves a better performance than all other methods. Moreover, all these approaches require an additional person detector either to get the bounding boxes \cite{Iqbal_ECCVw2016, wei2016convolutional}, or the rough scale of the persons \cite{insafutdinov2016deepercut}. Our approach on the other hand does not require a separate person detector, and we perform joint detection across different scales, while also solving the person association problem across frames. 

We also visualize how multi-person pose estimation accuracy (mAP) relates with the multi-person tracking accuracy (MOTA)  in Fig.~\ref{fig:MOTAvsMAP}. Finally, Tab.~\ref{tab:computational_analysis} provides mean and median runtimes for constructing and solving the spatio-temporal graph along with the graph size for $k\!=\!31$ frames over all test videos.  

\begin{table}[htbp!]
\vspace{-1mm}
\centering
%\scriptsize
\footnotesize
    \begin{tabularx}{\columnwidth}{@{}l *5{>{\centering\arraybackslash}X}@{}}
       \hline
      	&  Runtime (sec./frame)      & \# of nodes     	& \# of spatial edges  		& \# of temp. edges    \\ \hline
Mean  	& 14.7				 &			2084		&		65535		&  12903    		\\  
Median  &  4.2   			  &			1907		&		58164	 	&  8540					\\ \hline 
\end{tabularx}
\caption{Runtime and size of the spatio-temporal graph ($\tau\!=\!3$, HT:N:S, $k\!=\!31$), measured on a single threaded 3.3GHz CPU . \vspace{-7mm}
}
\label{tab:computational_analysis}
\end{table}

\section{Conclusion}
\vspace{-1mm}
In this paper we have presented a novel approach to simultaneously perform multi-person pose estimation and tracking. We demonstrate that the problem can be formulated as a spatio-temporal graph which can be efficiently optimized using integer linear programming. We have also presented a challenging and diverse annotated dataset with a comprehensive evaluation protocol to analyze the algorithms for multi-person pose estimation and tracking. Following the evaluation protocol, the proposed method does not make any assumptions about the number, size, or location of the persons, and can perform pose estimation and tracking in completely unconstrained videos. Moreover, the method is able to perform pose estimation and tracking under severe occlusion and truncation. Experimental results on the proposed dataset demonstrate that our method outperforms other baseline methods. \vspace{-7mm} \\ \\ 

\noindent\textbf{Acknowledgments.} The authors are thankful to Chau Minh Triet, Andreas Doering, and Zain Umer Javaid for the help with annotating the dataset. The work has been financially supported by the DFG project GA 1927/5-1 (DFG Research Unit FOR 2535 Anticipating Human Behavior) and the ERC Starting Grant ARCA (677650).

{\small   
\bibliographystyle{ieee}
\bibliography{pose_bib,refs-short,anton-ref}
}

\end{document}